\documentclass[sigconf]{acmart}
\usepackage{flushend}
\usepackage{graphicx}
\usepackage{ifthen}    
\usepackage{xcolor}
\usepackage{tcolorbox}
\usepackage{etoolbox}  
\usepackage{array}
\usepackage{makecell} 
\usepackage{booktabs}

\usepackage{multirow}
\usepackage{booktabs}
\usepackage{graphicx} 
\usepackage{subcaption} 
\usepackage{placeins} 
\usepackage{amssymb}
\usepackage{xspace}

\usepackage{adjustbox} 
\usepackage{algpseudocode}

\usepackage[ruled,vlined]{algorithm2e}
\usepackage{amsmath}
\usepackage{float}
\usepackage{colortbl}
\tcbuselibrary{breakable} 
\usepackage{tabularx} 
\usepackage{xstring}
\usepackage{cleveref}

\AtBeginDocument{%
  }

\copyrightyear{2025}
\acmYear{2025}
\setcopyright{cc}
\setcctype{by}
\acmConference[KDD '25]{Proceedings of the 31st ACM SIGKDD Conference on Knowledge Discovery and Data Mining V.2}{August 3--7, 2025}{Toronto, ON, Canada}
\acmBooktitle{Proceedings of the 31st ACM SIGKDD Conference on Knowledge Discovery and Data Mining V.2 (KDD '25), August 3--7, 2025, Toronto, ON, Canada}
\acmDOI{10.1145/3711896.3737011}
\acmISBN{979-8-4007-1454-2/2025/08}

\begin{document}

\title{\sys: Reasoning over Graph Paths for Drug Repurposing\\and Drug Interaction Prediction}

\author{Tassallah Abdullahi}
\email{tassallah_abdullahi@brown.edu}
\affiliation{%
  \institution{Brown University}
  \city{Providence}
  \state{RI}
  \country{USA}
}


\author{Ioanna Gemou\textsuperscript{\dag}}
\email{ioanna_gemou@brown.edu}
\affiliation{%
  \institution{Brown University}
  \city{Providence}
  \state{RI}
  \country{USA}
}

\author{Nihal V. Nayak}
\email{nihal_vivekanand_nayak@brown.edu}
\affiliation{%
  \institution{Brown University}
  \city{Providence}
  \state{RI}
  \country{USA}
}

\author{Ghulam Murtaza}
\email{ghulam_murtaza@brown.edu}
\affiliation{%
   \institution{Brown University}
  \city{Providence}
   \state{RI}
  \country{USA}
}

\author{Stephen H. Bach }
\email{stephen_bach@brown.edu}
\affiliation{%
  \institution{Brown University}
  \city{Providence}
  \state{RI}
  \country{USA}
}

\author{Carsten Eickhoff}
\authornote{co-corresponding authors. \\ \textsuperscript{\dag}Also affiliated with the Technical University of Denmark.}
\email{carsten.eickhoff@uni-tuebingen.de}
\affiliation{%
  \institution{University of Tübingen}
  \city{Tübingen}
  \country{Germany}
}

\author{Ritambhara Singh}
\authornotemark[1]
\email{ritambhara@brown.edu}
\affiliation{%
  \institution{Brown University}
  \city{Providence}
  \state{RI}
  \country{USA}
}
\renewcommand{\shortauthors}{Abdullahi et al.}


\newcommand{\sys}{K-Paths\xspace}
\newcommand{\nihal}[1]{{\textcolor{magenta}{[Nihal: #1]}}}
\newcommand{\amina}[1]{\textcolor{purple}{\small \bf [Amina\#\arabic{amina}\stepcounter{amina}: #1]}}
\newcommand{\ioanna}[1]{\textcolor{blue}{\small \bf [Ioanna\#\arabic{ioanna}\stepcounter{ioanna}: #1]}}

\begin{abstract}

Biomedical knowledge graphs (KGs) encode rich, structured information critical for drug discovery tasks, but extracting meaningful insights from large-scale KGs remains challenging due to their complex structure. Existing biomedical subgraph retrieval methods are tailored for graph neural networks (GNNs), limiting compatibility with other paradigms, including large language models (LLMs). We introduce \sys, a model-agnostic retrieval framework that extracts structured, diverse, and biologically meaningful multi-hop paths from dense biomedical KGs. These paths enable prediction of unobserved drug–drug and drug–disease interactions, including those involving entities not seen during training, thus supporting inductive reasoning. \sys is training-free and employs a diversity-aware adaptation of Yen’s algorithm to extract the K shortest loopless paths between entities in a query, prioritizing biologically relevant and relationally diverse connections. These paths serve as concise, interpretable reasoning chains that can be directly integrated with LLMs or GNNs to improve generalization, accuracy, and enable explainable inference. Experiments on benchmark datasets show that \sys improves zero-shot reasoning across state-of-the-art LLMs. For instance, Tx-Gemma 27B improves by 19.8 and 4.0 F1 points on interaction severity prediction and drug repurposing tasks, respectively. Llama 70B achieves gains of 8.5 and 6.2 points on the same tasks. \sys also boosts the training efficiency of EmerGNN, a state-of-the-art GNN, by reducing the KG size by 90\% while maintaining predictive performance. Beyond efficiency, \sys bridges the gap between KGs and LLMs, enabling scalable and explainable LLM-augmented scientific discovery. We release our code and the retrieved paths as a benchmark for inductive reasoning.\footnote{\url{https://github.com/rsinghlab/K-Paths}}

\end{abstract}

%



\maketitle

\section{Introduction}
Drug development and safety assessment have traditionally been time-intensive and costly, often spanning years and requiring billions of dollars \cite{paul2010improve, dimasi2003price}. 
Recent advances in computational power and the availability of clinical and biological data are transforming this landscape, enabling faster, more cost-effective discovery and validation of safer drugs \cite{vamathevan2019applications, huang2024foundation}.
For instance, deep learning models trained on genomic and chemical datasets now predict drug efficacy and toxicity with high accuracy, reducing the need for extensive lab experiments \cite{chaves2024txllm}. 
Yet, the sheer volume and heterogeneity of these datasets pose integration challenges, making it difficult to extract meaningful insights \cite{gligorijevic2015methods, davidson1995challenges}.

Knowledge Graphs (KGs) offer a structured solution by integrating complex biological relationships, linking diseases, drugs, and proteins into an interconnected framework \cite{hetionet2017, kuhn2007stitch}. Despite their potential, the scale and complexity of KGs make it difficult to retrieve task-related information efficiently \cite{peng2023knowledge}.
While typical KGs contain thousands of nodes and millions of edges, only a small subgraph is relevant for a given task \cite{yu2021sumgnn}, limiting practical utility in domains like drug discovery, where precise, targeted insights are critical. Graph Neural Networks (GNNs) have proven effective in leveraging KGs for drug discovery, excelling in link prediction and capturing biological relationships \cite{Schlichtkrull2017ModelingRD, zitnik2018modeling,huang2024foundation}.
However, GNNs incur high computational costs on large-scale KGs ~\cite{abdallah2024task} and critically fail to generalize to unseen entities \cite{hamilton2017inductive}, a drawback in drug discovery where new drugs and diseases continuously emerge.

Large Language Models (LLMs) offer a promising alternative, excelling in zero-shot reasoning and inductive generalization \cite{kojima2022large, abdullahi2024learning, abdullahi2024retrieval}. Recent works have shown that LLMs can be grounded in KGs to enhance factual accuracy and reduce hallucinations in specialized domains ~\cite{abdullahi2024retrieval, abdullahi2024retrieval_jmir,edge2024localglobalgraphrag,wu2024medical}. However, extracting meaningful insights from KGs using LLMs is an ongoing challenge ~\citep{fatemi2023talk, perozzi2024let}. 
Current approaches to KG integration often exhibit limitations. General-purpose methods like ~\citep{ luo2024reasoninggraphsfaithfulinterpretable, mavromatis2024gnnraggraphneuralretrieval, shu2024knowledge} focus on narrow tasks such as question answering or binary classification, using extraction methods that struggle to scale to dense biomedical KGs. Conversely, biomedical-specific approaches like~\citep{zhang2023emergingdruginteractionprediction} are tightly coupled to GNN architectures, hindering their direct integration with modern LLMs without substantial modifications. Addressing these limitations is crucial for developing a generalizable and scalable framework for drug
discovery.

\begin{figure}[t]
    \centering
    \includegraphics[width=\linewidth]{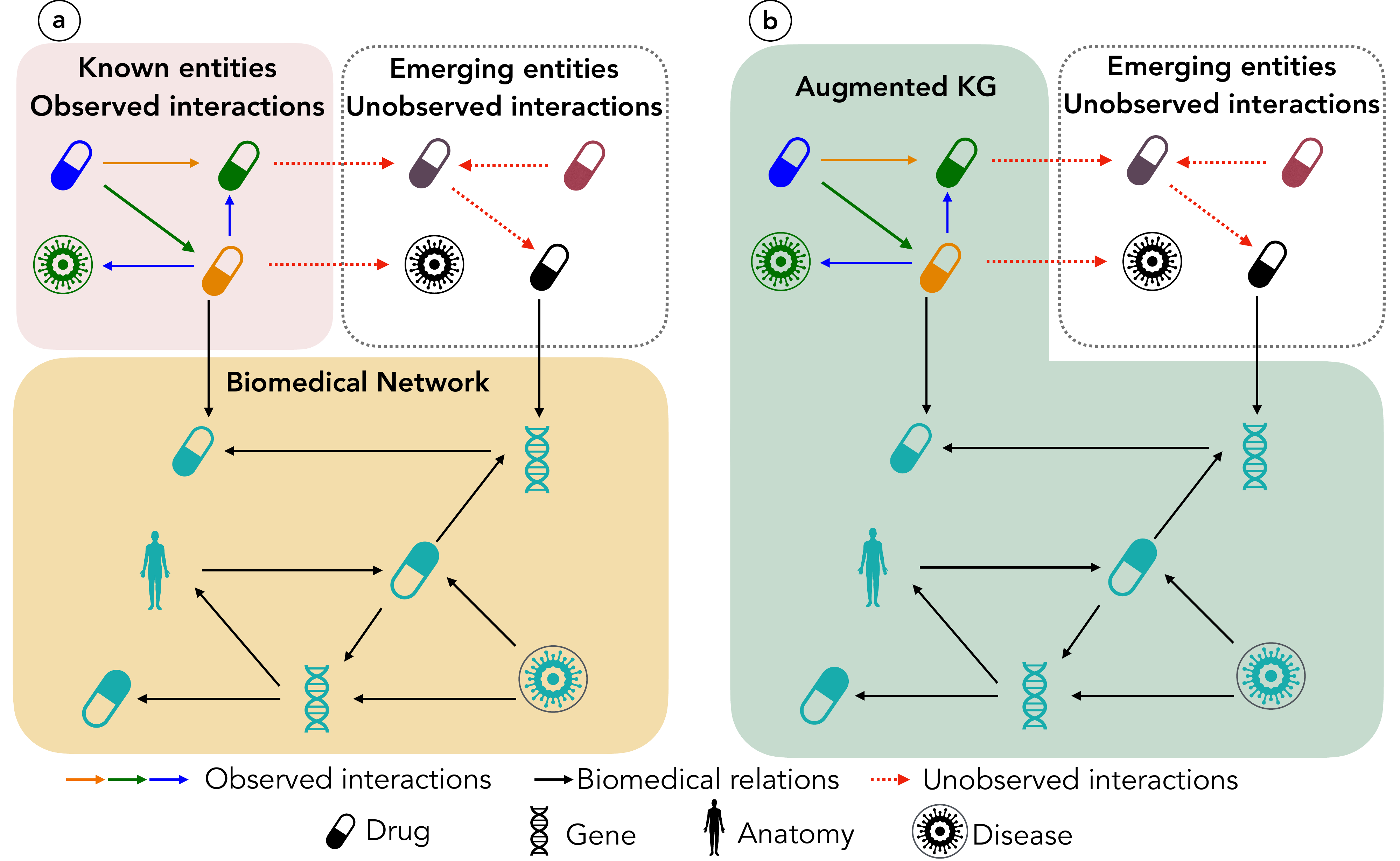} 
    \caption{\small Schematic representation of the predictive framework for unobserved interactions. (a) Problem formulation: Given a network of observed interactions among known entities (drugs or diseases) and a broader biomedical network containing additional relationships between various entities (e.g., drugs, diseases, genes, etc.). The task is to predict unobserved interactions between a known entity and an emerging entity or two emerging entities. (b) Augmented KG: The observed interactions and biomedical network are integrated to create a richer representation for the task.}
    \label{fig:overview}
\end{figure}

We introduce \textbf{``\sys''}, a novel model-agnostic retrieval framework designed to extract highly relevant entities and relationships from large biomedical KGs to aid in predicting unobserved drug–disease and drug–drug interactions, including those involving entities unseen during training. Unlike traditional approaches, \sys generates structured, interpretable multi-hop paths directly usable by LLMs, enabling efficient and accurate zero-shot reasoning.
\sys is particularly valuable in drug discovery, where identifying such interactions can facilitate drug repurposing and safer treatment opportunities.
\Cref{fig:overview} illustrates the problem \sys addresses: we aim to predict unobserved interactions between known and emerging entities (or between two emerging entities), where no interaction of interest has been observed previously. This formulation establishes an inductive reasoning setting, enabling predictions beyond observed edges.

\begin{figure*}[t]
    \centering
    \includegraphics[width=\linewidth]{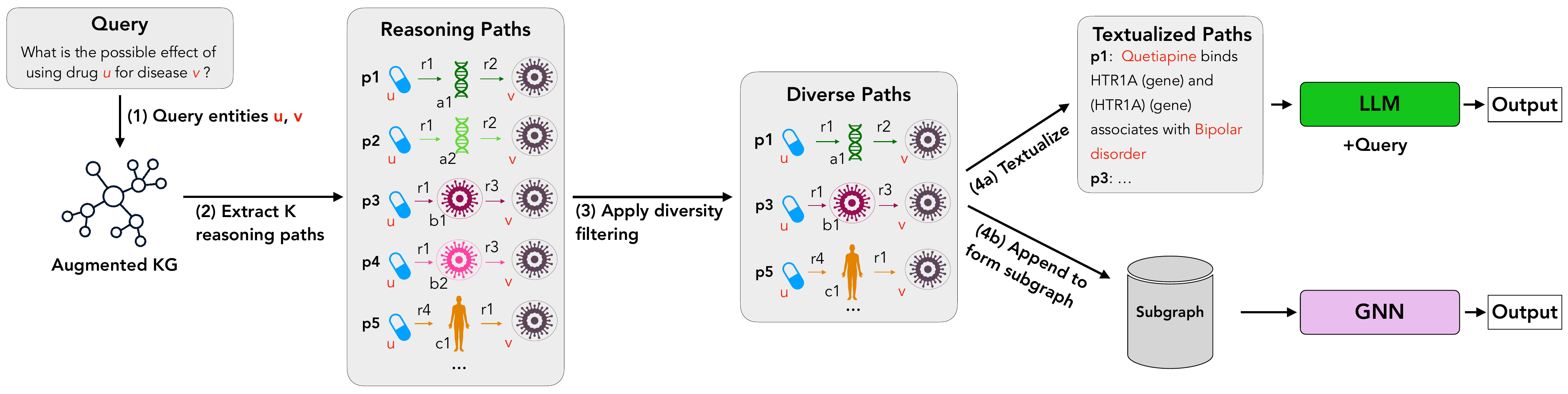} 
    \caption{\small \sys Overview. (1) Given a query about the effect of an entity ($u$) on another entity ($v$), (2) \sys extracts reasoning paths from an augmented KG connecting ($u$) and ($v$). (3) These paths are filtered for diversity and (4a) transformed into natural language descriptions for LLM inference. (4b) The retrieved paths can also be used to construct a subgraph, enabling GNNs to leverage more manageable information for training and prediction.}
    \label{fig:pipeline-overview}
\end{figure*}

\sys is training-free and operates in three steps as shown in \cref{fig:pipeline-overview}.
First, it augments the biomedical KG by integrating observed interactions from training data, enhancing its representation of interactions. 
This enriched KG serves as the network for retrieving biologically meaningful connections. 
Next, it retrieves biologically meaningful connections using our diversity-aware adaptation of Yen's algorithm \cite{yen1971finding}.
While Yen’s algorithm iteratively finds alternative shortest paths, our adaptation prioritizes relational diversity and biological relevance over redundant shortest-path variations. 
Finally, the retrieved paths are transformed into natural language representations, enabling LLMs to reason over them and effectively predict interactions. 
These paths can also be used to construct task-specific subgraphs, allowing GNN models to operate on a more focused graph and substantially reducing computational overhead.

We evaluate \sys in both zero-shot generative and supervised learning settings on drug repurposing and drug–drug interaction (DDI) tasks involving emerging entities.
Experiments show that \sys significantly enhances LLM reasoning in a zero-shot setting: on DDI severity prediction, Tx-Gemma 27B improves by 19.8 F1 points, and on drug repurposing tasks, by 4.0. Llama 70B achieves similar gains of 8.5 and 6.2 points on the same tasks, respectively. In supervised settings, \sys reduces KG size by 90\% and improves training efficiency for EmerGNN, a state-of-the-art GNN model, \cite{zhang2023emergingdruginteractionprediction}, without significant loss in performance. More importantly, the retrieved paths provide interpretable rationales for predicted interactions, enhancing explainability and offering valuable biological insights.

\section{Related work}
\newcommand{\heading}[1]{\smallskip\noindent\textbf{#1}\enspace}

\subsection{Biomedical KGs}
Biomedical KGs have been widely used to model the complex relationships among drugs, diseases, genes, and other biological entities, enabling data-driven advances in drug discovery~\citep{zitnik2018modeling, huang2024foundation, yu2021sumgnn, wang2024accurate}. 
These large-scale KGs are often curated for specialized purposes such as drug-disease interactions~\citep{wishart2018drugbank,hetionet2017}, gene-drug interactions~\citep{WhirlCarrillo2021}, and protein-protein interactions~\citep{gao2023hierarchical}.
In this work, we focus on drug-disease interactions and drug-drug interactions as they play a critical role in drug discovery, clinical decision-making, and patient safety.
A key limitation of biomedical KGs is that they can be incomplete in several ways.
They may lack observed interactions between existing entities (e.g., drugs or diseases) or fail to capture interactions involving emerging entities- those newly introduced or underrepresented in the KG.
We focus on predicting unobserved interactions among existing and emerging drugs or diseases, as well as interactions involving emerging drugs or diseases. 
This setting requires inductive reasoning, where a model must generalize to unseen nodes \citep{zhang2023emergingdruginteractionprediction}, as opposed to transductive reasoning, which assumes full knowledge of the graph at training time.
The inductive nature of this challenge reflects real-world biomedical discovery, where new drugs and diseases are continuously introduced.

\subsection{GNNs for drug discovery}
GNNs, including variants of graph convolutional networks, have demonstrated strong performance in modeling the structure of biomedical KGs and predicting unobserved interactions between entities~\citep{zhang2023emergingdruginteractionprediction, wang2024accurate, lin2020kgnn}.
These models often leverage external KGs such as Hetionet~\citep{hetionet2017} to learn richer representations of biomedical entities.
However, training on large-scale KGs is computationally expensive.
To address scalability, recent methods like \cite{wang2024accurate, yu2021sumgnn} extract fixed-sized subgraphs before passing them to the graph network.
While effective, these approaches are typically designed for transductive settings, where both nodes involved in a prediction are observed during training. This limits their ability to generalize to emerging entities, a critical need in drug discovery.
In this work, we operate in the inductive setting, aiming to predict interactions involving unseen drugs or diseases.

Our work is closely related to EmerGNN~\citep{zhang2023emergingdruginteractionprediction}, which augments the training network with Hetionet and uses a flow-based model to extract paths and construct subgraphs for GNN processing. 
However, EmerGNN requires a separately trained model and relies on beam search to generate paths tailored for graph architectures.
In contrast, \sys is training-free, retrieves diverse and biologically meaningful paths with minimal computational overhead, and supports model-agnostic integration with GNNs and LLMs. 
This flexibility enables \sys to support both supervised and zero-shot learning.
Furthermore, in \cref{sec:results}, we show that integrating \sys into EmerGNN improves efficiency and performance compared to using the entire KG. 
\sys also significantly improves LLM zero-shot performance without any parameter updates.

\subsection{LLMs for drug discovery}
LLMs have emerged as powerful tools in biomedical research, with applications ranging from conversational agents for drug discovery~\citep{wang2025txgemma}, and drug repurposing~\citep{inoue2024drugagent}, to molecular understanding~\citep{liang2023drugchat}. 
A key factor in their success is in-context learning~\citep{brown2020language}, which enables LLMs to incorporate new information, such as retrieved examples or structured facts, at inference time without fine-tuning.
This ability is particularly valuable for multi-hop reasoning, where models synthesize information across multiple steps or sources, such as KG paths, to derive accurate conclusions~\citep{mavromatis2024gnnraggraphneuralretrieval}. 
Recent studies show that LLMs can reason faithfully over KG paths to answer complex factual queries~\citep{mavromatis2024gnnraggraphneuralretrieval, luo2024reasoninggraphsfaithfulinterpretable}.
However, this potential remains underexplored in the context of biomedical KGs, particularly in inductive settings, where multi-hop reasoning could improve tasks like predicting interactions involving novel drugs or emerging diseases.

\paragraph{Graph-Augmented LLMs:}
\sys builds upon emerging research at the intersection of retrieval-augmented generation (RAG) and graph-based reasoning~\citep{luo2024reasoninggraphsfaithfulinterpretable, edge2024localglobalgraphrag}. Existing approaches construct synthetic KGs from unstructured text using LLMs \cite{edge2024localglobalgraphrag, wu2024medical} or rely on general-domain KGs like Freebase~\citep{luo2024reasoninggraphsfaithfulinterpretable} for question answering.
In contrast, \sys leverages curated biomedical KGs to enable precise, domain-specific inference.
Moreover, existing systems like RoG~\citep{luo2024reasoninggraphsfaithfulinterpretable} and GNN-RAG~\citep{mavromatis2024gnnraggraphneuralretrieval} require trained planning/retrieval modules or exhaustive path enumeration (e.g., KG-LLM's \cite{shu2024knowledge} depth-first search), which are computationally expensive and difficult to scale to dense KGs.
\sys circumvents these limitations by using a training-free, heuristic adaptation of Yen’s algorithm, with a filtering step for efficiency and relational diversity.

\paragraph{Design Trade-offs:}
RoG, GNN-RAG, and KG-LLM represent three distinct paradigms for path-based LLM reasoning, but each faces limitations for full-graph biomedical inference. 
RoG and GNN-RAG's retrieval modules operate on query-specific subgraphs generated via PageRank \cite{andersen2006local} and assume answers exist within the local subgraph.
This assumption may break down in inductive settings, where test-time entities may not appear during training. 
Additionally, their reliance on pre-generated subgraphs limits scalability to large biomedical KGs.
Similarly, KG-LLM is limited to a binary setup, and its exhaustive path enumeration over the full KG becomes infeasible in high-degree biomedical graphs.
In contrast, \sys enables scalable, explainable inference by retrieving diverse, biologically meaningful paths, without training or exhaustive search. We therefore analyze these approaches conceptually rather than as direct baselines, because they differ significantly in reasoning scope, scalability, and modeling assumptions.

\section{Approach} \label{sec:methods} 
\subsection{Problem definition}
Building on the problem illustrated in \cref{fig:overview}, we aim to predict unobserved interactions in an inductive reasoning setup, as described in \citet{zhang2023emergingdruginteractionprediction}. These interactions can involve:
\begin{itemize}
\item A known entity (e.g., a well-studied drug or disease with some observed interactions) and an emerging entity (e.g., a drug or disease whose interactions of interest have not been observed), or
\item Two emerging entities.
\end{itemize}
The unobserved interactions may include drug-drug interactions or drug-disease interactions.

Formally, we define a knowledge graph $\mathcal{G} = \{(u, r, v) \mid u, v \in \mathcal{E}, r \in \mathcal{R} \}$ where $u, v \in \mathcal{E}$ represent biomedical entities (e.g., drugs, diseases, genes) and $r \in \mathcal{R}$ denotes a relation type. These relation types include known drug-drug and drug-disease interactions (observed interactions) and a broader biomedical network with relationships like drug-gene or gene-gene interactions.

Given two query entities \( u\) and \( v\), our goal is to infer their interaction type, which is framed as predicting the presence and type of relation $r$ between $u$ and $v$.
We define a computational model \(\phi(\mathcal{G})\) to predict these interactions under both zero-shot generative and supervised settings.
Specifically, we leverage LLMs for reasoning-based inference and GNNs for interaction prediction. 

\subsection{\sys framework}
We introduce \textbf{\sys}, a framework for predicting the interaction between entities \( u \) and \( v \). \sys comprises three key components:

\begin{enumerate}
    \item An \textit{augmented KG}: This module constructs a knowledge graph $\mathcal{G}$ by integrating known drug-drug or drug-disease interactions with a broader biomedical KG. This integration incorporates additional entities such as genes and their known relationships (e.g., drug-gene or gene-gene interactions). 
    \item A \textit{diverse path retrieval module}: This module employs a novel path retrieval algorithm to retrieve a diverse set of relevant reasoning paths connecting the query entities \( u \) and \( v \) from the augmented KG $\mathcal{G}$.
    \item A \textit{path integration module}: This module processes the retrieved query-specific paths for interaction prediction.
    For LLM-based reasoning and interaction inference, the paths are transformed into natural language and appended to the interaction query prompt.
    For GNN-based interaction prediction, the paths are reconstructed into query-specific subgraphs.
\end{enumerate}
The overall \sys framework is illustrated in \cref{fig:pipeline-overview}.

\subsection{Augmented KG} \label{aug-network}
Following prior work~\citep{yu2021sumgnn, wang2024accurate, zhang2023emergingdruginteractionprediction}, we define the \textit{augmented KG} as our knowledge graph \(\mathcal{G}\), constructed by integrating:  
\begin{enumerate}
    \item Observed interactions (e.g., drug-drug and drug-disease) from the training set.
    \item Hetionet, a biomedical knowledge graph containing biological entities (e.g., genes, proteins, pathways) and their relationships (e.g., drug-gene, gene-gene, protein-protein interactions)~\citep{hetionet2017}.
\end{enumerate}


Since Hetionet is incomplete, incorporating known interactions from the training set enhances coverage and helps build a more comprehensive augmented KG. 
However, interactions involving emerging entities remain missing in this graph. To address this, we leverage existing relationships within the augmented KG to infer these unobserved interactions. Additionally, following~\citep{zhang2023emergingdruginteractionprediction}, we incorporate inverse relations to account for the directed nature of the augmented KG, ensuring bidirectional information flow.

The augmented KG \(\mathcal{G}\) serves as the structured knowledge graph for all subsequent tasks.

\begin{algorithm}[!ht]
\KwIn{A set of paths $P = \{p_1, p_2, \dots, p_K\}$, 
      where each path $p$ is described by \\
      $R(p) = (r_1, r_2, \dots, r_l)$ (sequence of relations) and \\
      $\mathcal{E}(p) = (e_1, e_2, \dots, e_m)$ (sequence of entities).}
\KwOut{A subset $P' \subseteq P$ with redundant paths removed.}

\BlankLine
Initialize $P' \leftarrow \emptyset$\;

\ForEach{path $p \in P$}{
  Let $R(p) = (r_1, r_2, \dots, r_l)$ be the relation sequence of $p$\;
  Let $l = |R(p)|$ (the length of the relation sequence)\;

  \If{\(\nexists\, q \in P'\) such that \(R(q) = R(p)\) 
      \textbf{and} \(|R(q)| = l\)}{
    $P' \leftarrow P' \cup \{p\}$\;
  }
}
\Return $P'$
\caption{Filtering algorithm to remove relationally redundant retrieved paths.}
\label{alg:filtering}
\end{algorithm}

\subsection{Diverse path retrieval module}
The path retrieval module is a key component of our framework.
It provides the downstream computational model with highly relevant yet manageable information from our augmented KG $\mathcal{G}$.

Our diverse path retrieval algorithm retrieves a set of $K$ shortest diverse paths between two entities, \( \{u, v\} \), from $\mathcal{G}$ using Yen's algorithm ~\cite {yen1971finding}.
We prioritize shortest paths for several reasons. 
Shorter paths capture stronger, more interpretable relationships, while longer paths introduce noise and uncertainty \cite{liben2003link, barabasi2004network}. 
Finally, empirical studies on biomedical KGs further show that meaningful interactions typically occur within a few hops\cite{zitnik2018modeling, himmelstein2016pharmacotherapydb}.

Yen's algorithm extends Dijkstra's algorithm \cite{dijkstra1959note} by iteratively computing shortest paths while temporarily excluding specific edges, thus generating progressively longer, loop-free alternatives.
The output from this process contains relationally redundant paths. 
Therefore, we introduce a filtering algorithm (\cref{alg:filtering}) to remove these redundancies.
Our filtering algorithm eliminates paths with duplicate relation sequences of the same path length.
This filtering step results in a diverse set of paths, which are passed to the next module in our framework. 

\textbf{Edge cases: }In cases where only one of \( \{u, v\} \) exists in $\mathcal{G}$, we retrieve its immediate neighbors and their connecting relationships, applying the same filtering algorithm.  
This provides valuable context even when a complete path between two entities cannot be established.  
If neither entity exists in $\mathcal{G}$, no information is retrieved.

\subsection{Path integration module}
This module integrates the retrieved diverse paths to predict interactions between the query entities.
We explore two distinct approaches: LLM-based reasoning and GNN-based prediction.

\heading{LLM reasoning} \label{llm-reasoning}
We convert the entities and relations from the diverse paths into natural language using predefined dictionaries.
These dictionaries map entities and relations to their respective types and textual representations.
To improve clarity, we append entity type descriptors in parentheses after each entity.

For example, as illustrated in \cref{fig:pipeline-overview}, consider the path:
\[
p_1 : u \rightarrow r_1 \rightarrow a_1 \rightarrow r_2 \rightarrow v
\]
representing the relationship between \textit{Quetiapine} and \textit{Bipolar disorder}. This path is transformed into the natural language description:

\begin{quote}
    \textit{Quetiapine binds HTR1A (gene) and HTR1A (gene) associates with Bipolar disorder}.
\end{quote}

Following \cite {zhang2023emergingdruginteractionprediction}, if the retrieved relation belongs to an inverse relation category, we convert it into passive voice. For instance, the relation:

\begin{quote}
    \textit{(Disease) downregulates (Gene)}
\end{quote}

is converted into:

\begin{quote}
    \textit{(Gene) is downregulated by (Disease)}.
\end{quote}

This explicit type mapping helps the LLM understand the semantic roles of each entity, even if it is unfamiliar with domain-specific entities like \textit{``HTR1A''}.
This conversion process is applied to all \textit{K} retrieved paths.
In the zero-shot setting, these textualized paths are appended to the original query, providing contextual information for the LLM to perform inference.
In the supervised setting, we can fine-tune the LLM using these textualized paths as training data.
Furthermore, representing the retrieved paths in natural language enhances the explainability of the LLM's predictions.

\heading{GNN interaction prediction} \label{gnn-learning}
For GNN-based prediction, instead of directly inputting the large augmented KG into the GNN, we use the diverse paths to construct smaller, query-specific subgraphs.
This approach significantly reduces the computational complexity and allows GNN to focus on the most relevant information.
During training, the GNN learns entity-specific representation by aggregating information from known drug-drug or drug-disease interactions and their corresponding query-specific subgraphs. 
These learned representations are then used to predict the type of interaction between entity pairs.  
During testing, we extend the learned representation space by incorporating new test nodes and their corresponding query-specific subgraphs.
This allows us to evaluate the model's ability to accurately predict unobserved interactions using this extended graph in a supervised inductive setting.

\section{Experiments} \label{sec:experiments}
In this section, we aim to answer the following research questions:
\textbf{RQ1:} Can LLMs accurately predict interaction \textit{types}, \textit{severities}, or \textit{indications} by reasoning over multi-hop knowledge graph paths provided as context in a zero-shot setting? 

\noindent
\textbf{RQ2:} How do path selection strategies (e.g., shortest path, diverse path selection) and the number of reasoning paths \textit{(K)} influence LLM performance in interaction prediction tasks? 

\noindent
\textbf{RQ3:} How do subgraphs derived from the \sys framework impact the performance of LLMs and GNNs in a supervised setting?

\begin{table}[H] 
\caption{\small Datasets Overview. \textit{u} and \textit{v} are drugs or disease entities.}
\centering
\scriptsize
\setlength{\tabcolsep}{2pt}      
\renewcommand{\arraystretch}{1}  
\begin{tabularx}{\columnwidth}{l c c X}
  \toprule
  \textbf{Dataset} & \textbf{Entities} & \textbf{Categories} & \textbf{Example} \\
  \midrule
  DDInter            & 1689 drugs                     & 3 severity levels          & Severity: Major (\textit{u} + \textit{v}) \\
 
  DrugBank           & 1710 drugs                     & 86 interaction levels      & \textit{u} decreases \textit{v's} excretion rate\\
 PharmacotherapyDB  & 601 drugs; 97 diseases           & 3 indications              & \textit{u} treats \textit{v} \\
  
  \bottomrule
\end{tabularx}
\label{tab:tasks_overview}
\end{table}

\subsection{Datasets} \label{sec:datasets}  
\heading {Evaluation datasets:} Following previous work, we conducted experiments in an inductive setting to assess the model's generalization to unseen entities.
In this setting, the training set includes interactions only between entities in the training set, the validation set contains interactions where at least one entity appears only in the validation set or interactions exclusively within it, and the test set follows the same rule for test entities.
This setup enables evaluation of the model's ability to predict unobserved interactions involving both known and emerging entities, as well as interactions between emerging entities.

We used three datasets with different task objectives (Table~\ref{tab:tasks_overview}): DDInter \cite{xiong2022ddinter}, PharmacotherapyDB (v1.0) \cite{himmelstein2016pharmacotherapydb}, and DrugBank \cite{wishart2018drugbank}.
For DrugBank, we applied the inductive split from \cite{du2024customized}, while for DDInter, we created train, validation, and test sets following the described inductive split. Due to its smaller size, PharmacotherapyDB was divided into training and test sets in a similar inductive setting. 
Table 1 provides detailed statistics, including the number of specific interaction types.

\heading {External knowledge base:} Hetionet is a heterogeneous biomedical network curated from 29 databases. It includes various biomedical entities such as compounds, genes, diseases, etc. Following \cite{yu2021sumgnn}, the processed version used in this work includes 33,765 nodes across 11 types and 1,690,693 edges spanning 23 relation types.

\subsection{Baselines}
We compare the \sys framework  with the following baselines:

\subsubsection{LLM-based baselines:}
Here, we evaluated \sys' impact on LLM reasoning in zero-shot and supervised settings. 
\begin{itemize}
    \item \textbf{Reasoning based on internal knowledge (Base):} The LLM uses only its internalized knowledge to infer interactions and relationships.
    \item \textbf{Reasoning based on textual definitions (Definitions):} The LLM leverages textual definitions of drugs and diseases from external resources, such as DrugBank\footnote{\url{https://go.drugbank.com/}} and the Disease Ontology\footnote{\url{https://disease-ontology.org/}}, to infer interactions and relationships independent of the knowledge graph's structure.

\end{itemize}

\subsubsection{GNN-based baselines:}
Here, we evaluated \sys' impact on GNNs in a supervised setting only.
\begin{itemize}
    \item \textbf{Complete graph-based prediction (Complete KG):} The Graph Neural Network (GNN) uses the entire augmented KG to predict interactions, considering all entities and relationships.
\end{itemize}

\subsection{Implementation details}
We provide the details for \sys implementation for all our experiments with different LLMs and GNNs. 
For further details on prompting strategies, training configurations, and hyperparameters, see \cref{app:prompting}.
During path retrieval, we limit the maximum path length to 3 as longer paths tend to be more susceptible to noise and less likely to represent meaningful, interpretable relationships.
We assume there is no direct link between the query entities whose interaction we aim to predict, and the model must infer the interaction type by reasoning over existing facts in the KG. 
To prevent data leakage, we explicitly check and remove direct interaction links between entities during training and verify their absence in the test set.
For all experiments across datasets, we set  \(K = 10\) retrieved paths per query.
As $K$ increases, both the number of hops and average path length grow accordingly.

For LLM-based reasoning, we employ a suite of models \textbf{LLaMA-3.1-8B-Instruct} and \textbf{LLaMA-3.1-70B-Instruct}~\cite{dubey2024llama}, \textbf{Tx-Gemma-9B-Chat} and \textbf{Tx-Gemma-27B-Chat}~\cite{wang2025txgemma}, as well as \textbf{Qwen2.5-14B-Instruct} and \textbf{Qwen2.5-32B-Instruct}~\cite{yang2025qwen3} as our primary inference models, leveraging their instruction-following capabilities.
In zero-shot settings, we utilize direct prompting to evaluate their reasoning ability.
In supervised settings, we fine-tuned Llama-3.1-8B-Instruct using QLoRA, a lightweight and efficient fine-tuning approach.
In both experiments, we generate the output with greedy decoding.
For GNN architectures, we use Relational Graph Convolutional Networks (RGCN) because of their ability to handle relational data, which aligns with our task. Additionally, we adopt EmerGNN’s backbone architecture, as it is the current state-of-the-art inductive graph-based model for drug-drug interaction prediction tasks.


    
    
    
\subsection{Evaluation}
Following prior work~\citep{zhang2023emergingdruginteractionprediction}, we evaluate all GNN and LLM models on a multi-class classification task using accuracy, macro-averaged F1-score, and Cohen's Kappa.
Since LLMs produce open-ended text responses, we first parse each output using regular expressions and map it to one of the predefined dataset labels. If a response cannot be reliably mapped to a valid label, we default the prediction to the majority class from the training set.
For the DrugBank dataset, which contains 86 long, descriptive label phrases, we supplement regex-based evaluation with BERTScore to capture semantic similarity and provide a more robust assessment beyond exact string matching.

\section{Results} \label{sec:results}
\subsection{\sys improves zero-shot reasoning} \label{result-zero-shot}
We evaluate LLMs' reasoning capabilities for drug-disease interaction prediction in three zero-shot scenarios with varying context: (1) \textbf{Base}---the LLM relies solely on its internal knowledge; (2) \textbf{Definitions}---textual definitions of drugs and diseases are provided; and (3) \textbf{\sys}---the LLM is provided with diverse paths retrieved by our \sys framework.
\Cref{tab:zero_shot_results} presents results across multiple LLM sizes and domains on the DDInter, PharmacotherapyDB, and DrugBank datasets.

\textbf{Consistent improvements with \sys:}
Across all datasets, LLMs significantly benefit from \sys, outperforming both Base and Definitions settings. 
On DDInter, \sys boosts the F1-score by 13.42 for Llama 3.1 8B and 6.18 for Llama 3.1 70B, compared to the Definitions setting.
Similar gains appear on PharmacotherapyDB with improvements of 12.45 (8B) and 8.46 (70B).
The Tx-Gemma models (9B/27B), specifically fine-tuned for therapeutic tasks, show remarkable gains with \sys---achieving 29.2 and 19.8 point F1-score improvements on DDInter, respectively.
In some cases, Definitions slightly improve performance over Base or even degrade it, but they consistently fall short of the substantial gains achieved with \sys.
This gap likely stems from the nature of the information provided.
\sys offers structured, contextualized knowledge, showing how entities are related.
Definitions, while informative, provide declarative knowledge that lacks crucial relational context, needed for LLMs to reason about interactions effectively.

\textbf{The harder the task, the larger the gap:}
DrugBank involves 86 interaction types. In the Base setting, most models hover around 1\% F1 or 19\% BERTScore, whereas with \sys they achieve 30–40\%, emphasizing the importance of structured external knowledge in such complex tasks.

\textbf{\sys eases need for scale \& domain-specific models:}
Larger models generally outperform smaller ones in the Base setting (e.g., 70B vs.\ 8B Llama), but these gains diminish with external knowledge. 
Surprisingly, on DDInter, smaller models (8B Llama, Qwen 14B) occasionally outperform their larger counterparts when using \sys, suggesting high-quality external knowledge can reduce reliance on model scale.
Likewise, Tx-Gemma’s domain-specific pre-training only helps consistently on PharmacotherapyDB. However, their performance varies across datasets, suggesting domain specialization doesn't guarantee optimal reasoning.
These results demonstrate that structured reasoning paths provide benefits for zero-shot interaction prediction, often outweighing the benefits of model scale or domain-specific pretraining.

\begin{table*}[t!]
    \caption{\small Performance comparison of different models on zero-shot reasoning tasks. \textbf{Bold} indicates the best performance, while \underline{underlined} denotes the second-best performance. \sys improves domain-specific reasoning in a zero-shot setting.}
    \centering
    \begin{adjustbox}{max width=0.9\textwidth}
    \begin{tabular}{llccc|ccc|cccc}
        \toprule
        \multirow{2}{*}{\textbf{Model}} & \multirow{2}{*}{\textbf{Setting}} & \multicolumn{3}{c|}{\textbf{DDInter}} & \multicolumn{3}{c|}{\textbf{PharmacotherapyDB}} & \multicolumn{3}{c}{\textbf{DrugBank}} \\
        \cmidrule(lr){3-5} \cmidrule(lr){6-8} \cmidrule(lr){9-12}
        & & Accuracy & F1 & Kappa & Accuracy & F1 & Kappa & Accuracy & F1 & Kappa & BERTScore \\
        \midrule
        \multirow{4}{*}{Llama 3.1 8B Instruct} 
        & Base            & 69.09  & 33.34  & 4.37   & \underline{57.94}  & \underline{51.91}  & \underline{35.18}  & 31.36  & 0.65   & 0.87 & \underline{18.29} \\
            & Defintions    & \underline{70.43}  & \underline{33.46}  & \underline{5.51}   & 57.54  & 51.70  & 34.38  & \underline{31.63}  & \underline{0.68}   & \underline{0.95} & 12.09 \\
        & \sys              & \textbf{75.93}  & \textbf{46.76}  & \textbf{36.99}  & \textbf{69.44} & \textbf{64.36}  & \textbf{51.34}  & \textbf{55.54}  & \textbf{40.46}  & \textbf{45.58} & \textbf{63.51} \\

        \midrule
        \multirow{4}{*}{Llama 3.1 70B Instruct} 
        & Base            & 70.51  & \underline{40.01}  & \underline{19.07}  & 60.71  & 59.00  & 39.38  & \underline{31.32}  & \underline{1.35}   & \underline{3.88} & 18.65 \\
        & Definitions   & \underline{71.61}  & 37.07  & 12.98  & \underline{62.30}  & \underline{61.14}  & \underline{42.30}  & 30.48  & 1.03   & 1.84 & \underline{19.69} \\ 
        & \sys & \textbf{78.01}  & \textbf{46.19} & \textbf{35.33}  & \textbf{71.03}  & \textbf{67.46}  & \textbf{54.66}  & \textbf{57.72}  & \textbf{46.91}  & \textbf{49.19} & \textbf{66.52} \\


        \midrule
    \multirow{4}{*}{Tx-Gemma-9B-chat} 
    & Base            & \underline{16.73} & \underline{16.75} & \underline{0.85} & \underline{67.86} & \underline{65.40} & \underline{49.08} & \underline{31.43} & \underline{0.85} & \underline{0.59} & 12.42 \\
    & Defintions    & 8.54  & 8.48  & 0.56 & 64.29 & 61.46 & 45.10 & 31.04 & 0.80 & \underline{0.59} & \underline{18.83} \\
    & \sys              & \textbf{56.73} & \textbf{45.94} & \textbf{26.62} & \textbf{70.63} & \textbf{66.04} & \textbf{53.09} & \textbf{53.38} & \textbf{41.05} & \textbf{44.64} & \textbf{63.14} \\

   \midrule
    \multirow{4}{*}{Tx-Gemma-27B-chat} 
    & Base            & \underline{27.52}& \underline{27.28} & \underline{4.83} & \underline{67.46} & \underline{67.49} & \underline{51.39} & 31.40 & \underline{0.79} & \underline{1.73} & 18.49 \\
    & Definitions    & 20.86  & 21.72  & 3.3 & 59.52 & 58.98 & 40.20 & \underline{31.47} & 0.77 & 1.6 & \underline{19.21} \\
    & \sys              & \textbf{60.88} & \textbf{47.06} & \textbf{26.56} & \textbf{72.62} & \textbf{71.49} & \textbf{58.19} & \textbf{50.71} & \textbf{43.93} & \textbf{40.33} & \textbf{58.78} \\

       \midrule
    \multirow{4}{*}{Qwen2.5-14B-Instruct} 
    & Base            & 19.76 & \underline{18.36} & \underline{0.78} & \underline{58.33} & \underline{57.91} & \underline{39.72} & 31.53 & \underline{0.61} & \underline{0.07} & \underline{5.00} \\
    & Definitions    & \underline{22.16}	& 18.23 &	0.43 & 57.94	& 57.3	& 39.13 & \underline{31.54}	& \underline{0.61}	& \underline{0.07} & -1.00 \\
    & \sys              &\textbf{66.25} &	\textbf{49.16}	& \textbf{30.03} & \textbf{65.08}	& \textbf{58.18}	& \textbf{45.75}  & \textbf{49.12} & \textbf{41.76} & \textbf{33.84} & \textbf{47.39} \\

    \midrule
    \multirow{4}{*}{Qwen2.5-32B-Instruct} 
    & Base           & \underline{38.23} & \underline{31.13} & \underline{2.6} & \underline{65.08} & \underline{65.03} & \underline{47.50} & \underline{31.95} &	\underline{0.78} &	\underline{1.85} & \underline{12.98} \\
    & Definitions    & 23.61	& 20.98 &	0.81 & 62.30	& 62.55	& 44.14 & 31.54	& 0.61	& 0.04 & 1.00 \\
    & \sys              &\textbf{63.69} &	\textbf{48.88}	& \textbf{29.23} & \textbf{71.83}	&\textbf{ 67.83}	& \textbf{55.10}  & \textbf{41.23} & \textbf{31.31} &	\textbf{20.02} & \textbf{32.69} \\ 

        \bottomrule
    \end{tabular}
    \end{adjustbox}
    \label{tab:zero_shot_results}

\end{table*}

\begin{table*}
    \caption{ \small Comparison of LLM responses based on external knowledge type. \sys allows for explainable inference.}
    \centering
    \scriptsize 
    \renewcommand{\arraystretch}{1.2} 
    \begin{tabular}{p{1.5cm} | p{7cm} | p{7cm}} 
    \toprule
        & \textbf{PharmacotherapyDB} & \textbf{DDinter} \\
        \midrule
        \textbf{Query} & \begin{minipage}[t]{6.5cm}Determine the possible effect of using \textbf{\textit{Vincristine}} (Drug) for \textbf{\textit{Muscle cancer}} (Disease).\end{minipage} & \begin{minipage}[t]{6.5cm}Determine the severity of interaction when \textbf{\textit{Ritonavir}} (Drug 1) and  \textbf{\textit{Leflunomide}} (Drug 2) are taken together.\end{minipage} \\
        \midrule
        \textbf{Answer} & \begin{minipage}[t]{6.5cm}Disease Modifying\end{minipage} & \begin{minipage}[t]{6.5cm}Major\end{minipage} \\
        \midrule
        \textbf{Definitions} & 
        \begin{minipage}[t]{6.5cm}
        \textbf{\textit{Vincristine}} is an antitumor that treats leukemia, lymphoma, Hodgkin's disease, and other blood disorders.\par
        \textbf{\textit{Muscle cancer}} is a musculoskeletal system cancer located in the muscle.
        \end{minipage} & 
        \begin{minipage}[t]{6.5cm}
        \textbf{\textit{Ritonavir}}, an HIV protease inhibitor, boosts other protease inhibitors' effectiveness and is used in some HCV therapies. As a CYP3A inhibitor, it increases drug concentrations. \par
        \textbf{\textit{Leflunomide}} is a pyrimidine synthesis inhibitor belonging to the disease-modifying antirheumatic drugs chemically and pharmacologically very heterogeneous.
        \end{minipage} \\
        \midrule
        \textbf{\sys} &  
        \begin{minipage}[t]{6.5cm}
        \textbf{\textit{Vincristine}} treats Kidney Cancer (Disease) and Kidney Cancer (Disease) resembles \textbf{\textit{Muscle Cancer}}.\par  
        \textbf{\textit{Vincristine}} downregulates \textit{TP53} (Gene) and \textit{TP53} (Gene) is associated with Muscle Cancer.
        \end{minipage} & 
        \begin{minipage}[t]{6.5cm}
        \textbf{\textit{Ritonavir}} binds CYP2C9 (Gene) and CYP2C9 (Gene) is bound by \textbf{\textit{Leflunomide}}\par
        \textbf{\textit{Ritonavir}} causes Neutropenia (Side Effect), and Neutropenia (Side Effect) is caused by \textbf{\textit{Leflunomide}}
        \end{minipage} \\
        \midrule
        \textbf{LLM Only} & \begin{minipage}[t]{6.5cm}Non Indications\end{minipage} & \begin{minipage}[t]{6.5cm}Moderate\end{minipage} \\
        \midrule
        
        \textbf{LLM+Definitions} & \begin{minipage}[t]{6.5cm}Non Indications\end{minipage} & \begin{minipage}[t]{6.5cm}Moderate\end{minipage} \\
        \midrule
        
        \textbf{LLM+\sys } & \begin{minipage}[t]{6.5cm}Disease Modifying\end{minipage} & \begin{minipage}[t]{6.5cm}Major\end{minipage} \\
        
        \bottomrule
    \end{tabular}
    \label{tab:llm_kg_comparison}
\end{table*}

\subsection{\sys enables explainable inference}
To complement our quantitative evaluation in \cref{result-zero-shot}, we conducted a qualitative analysis examining how the different forms of external knowledge—definitions and \sys influence the Llama 3.1 70B's responses. 
\Cref{tab:llm_kg_comparison} presents two case studies from PharmacotherapyDB and DDInter, comparing the model's predictions under different knowledge augmentation conditions.
We make two key observations.
(1) The Base model frequently predicts incorrectly, demonstrating insufficient domain knowledge.
In the PharmacotherapyDB example, although the Definition of \textit{Vincristine} mentions its use in cancer treatments, the model failed to infer its applicability to \textit{Muscle Cancer}.
However, incorporating reasoning paths from \sys corrects the LLM's predictions.
This suggests that factual definitions without explicit relational information are insufficient for correct reasoning in interaction prediction. 

(2)\sys enables the LLM to connect relevant entities and their contextual relationships, leading to improved accuracy. 
For instance, in the DDInter example, \sys allows the model to correctly predict a ``Major'' interaction between \textit{Ritonavir} and \textit{Leflunomide}, highlighting \textit{Neutropenia} as a possible side effect. This is crucial, as \textit{Neutropenia} is a potentially life-threatening condition.
A ``Moderate'' misclassification could have severe safety implications, stressing the importance of high-quality reasoning paths in safety-critical applications.
In summary, while Definitions provide some information about query entities, structured, diverse KG paths are essential for effective reasoning and improved zero-shot performance.
We further examine edge cases with limited or no KG paths in \cref{sec:extra-qualitative}, demonstrating \sys's graceful fallback to parametric knowledge when structural information is unavailable.

\begin{figure*}[htbp]
  \centering
  \begin{minipage}{0.4\textwidth} 
    \centering
    \includegraphics[width=\linewidth]{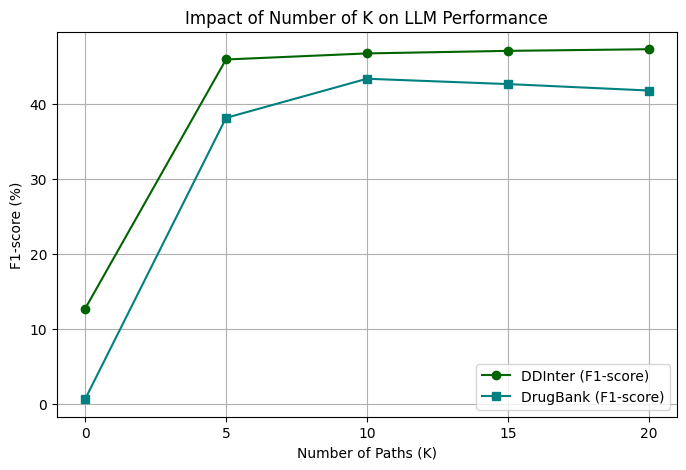}
  \end{minipage}
  \hspace{0.05\textwidth} 
  \begin{minipage}{0.4\textwidth}
    \centering
    \includegraphics[width=\linewidth]{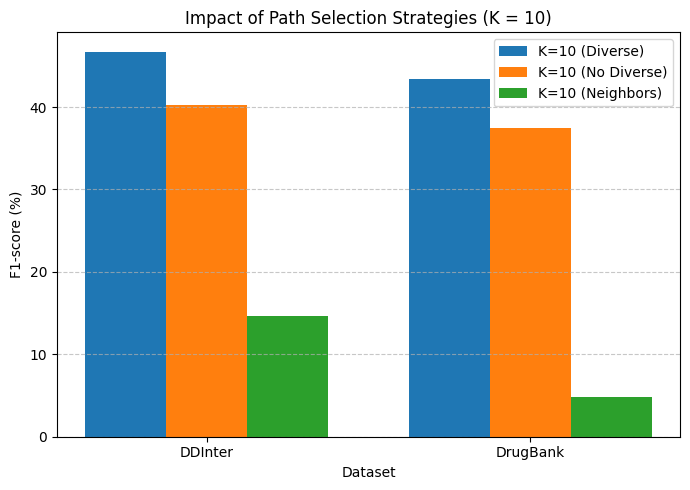}
  \end{minipage}
  
  \caption{\small Influence of path selection strategies on Llama 3.1 8B. Diverse paths are essential for performance improvement.}
  \label{fig:RQ2-results}
\end{figure*}

\begin{table*}[t]
    \caption{\small Supervised performance of various models across datasets. Bold indicates the best result within each model category; * marks the overall best performance.  With supervision, LLMs benefit from both knowledge types, while \sys enhances GNN efficiency without significant performance loss.}
    \centering
    \begin{adjustbox}{max width=0.7\textwidth}
    \begin{tabular}{llccc|ccc|ccc}
        \toprule
        \multirow{2}{*}{\textbf{Model}} & \multirow{2}{*}{\textbf{Setting}} & \multicolumn{3}{c|}{\textbf{DDInter}} & \multicolumn{3}{c|}{\textbf{PharmacotherapyDB}} & \multicolumn{3}{c}{\textbf{DrugBank}} \\
        \cmidrule(lr){3-5} \cmidrule(lr){6-8} \cmidrule(lr){9-11}
        & & Accuracy & F1 & Kappa & Accuracy & F1 & Kappa & Accuracy & F1 & Kappa \\
        \midrule
        
        \multicolumn{11}{l}{\textbf{LLM-Based Models}} \\
        QLoRA-Llama & Definitions & \textbf{82.36} & 67.96 & 53.64 & \textbf{81.75}\textsuperscript{*} & \textbf{79.91}\textsuperscript{*} & \textbf{71.84}\textsuperscript{*} & \textbf{73.22}\textsuperscript{*} & \textbf{68.15}\textsuperscript{*} & \textbf{67.26}\textsuperscript{*} \\
        
        QLoRA-Llama & \sys & 80.55 & \textbf{68.63} & \textbf{57.85} & 78.57 & 76.71 & 66.68 & 71.83 & 65.57 & 65.63 \\
        
        \midrule
        
        \multicolumn{11}{l}{\textbf{Graph-Based Models (GNNs)}} \\
        EmerGNN & Complete KG & 84.26 & 68.00 & \textbf{58.92}\textsuperscript{*} & \textbf{71.43} & \textbf{68.41} & \textbf{55.38} & \textbf{71.04} & \textbf{59.42} & \textbf{65.14} \\
        
        EmerGNN & \sys & \textbf{84.53}\textsuperscript{*} & \textbf{68.85}\textsuperscript{*} & 58.91 & 71.03 & 68.12 & 54.10 & 68.98 & 59.06 & 62.54 \\

        \midrule
        RGCN & Complete KG & 72.01 & 51.47 & 31.38 & 61.11 & 60.08 & \textbf{41.12} & 29.98 & 15.49 & 20.88 \\
            RGCN & \sys & \textbf{73.32} & \textbf{52.12} & \textbf{32.70} & \textbf{66.82} & \textbf{61.74} & 39.25 & \textbf{31.70} & \textbf{17.51} & \textbf{23.14 }\\
        \bottomrule
    \end{tabular}
    \end{adjustbox}
    \label{tab:results-3}
\end{table*}

\subsection{Path selection strategies and performance}
We study how number of retrieved paths ($K$) and path selection strategy affect the prediction performance of Llama 3.1 8B on the validation set.
We evaluate several path selection strategies: (1) the ``Base'' setting ($K=0$), (2) diverse paths from \sys($K=1, 5, 10, 15, 20$ ). As $K$ increases, the number of hops and path length increases. (3) shortest paths (without diversity filtering), and (4) local neighborhood edges (5 neighbors per entity). 
Results in \cref{fig:RQ2-results} show that adding diverse reasoning paths from \sys substantially improve performance.
On DDInter, F1-scores increase from 12.66\% ($K=0$) to 46.73\% ($K=10$), and on DrugBank, they increase from 0.59\% ($K=0$) to 43.35\% ($K=10$).  
However, performance gains diminish beyond $K=10$, suggesting that excessively long paths introduce redundancy or noise.  
\sys consistently outperforms shortest-path selection (without diversity filtering) and neighborhood selection.  
Omitting diversity filtering results in degradation of the F1-score by 6.99\% on DDInter and 4.32\% on DrugBank, highlighting the importance of diverse path retrieval.

\begin{table*}[t]
    \caption{\small Augmented KG Statistics: \sys Improves Training Efficiency.}
    \centering
    \begin{adjustbox}{max width=0.9\textwidth}
    \begin{tabular}{lccccccccc}
        \toprule
        \multirow{2}{*}{Dataset} & \multicolumn{4}{c}{Before Retrieval (Complete KG)} & \multicolumn{4}{c}{After Retrieval (\sys)} \\
        \cmidrule(lr){2-5} \cmidrule(lr){6-10}
         & \#Nodes & \#Relations & \#Triplets & Sec/Epoch \{EmerGNN, RGCN\} & \#Nodes & \#Relations & \#Triplets & Sec/Epoch \{EmerGNN, RGCN\} & Time (min) \\
        \midrule
        \textbf{DDInter}  & 35,107 & 26 & 1,763,596 & \{606.24s, 5.78s\} & 4,723 & 18 & 113,933 & \{102.21s, 1.18s\} & 13.57 \\
        \textbf{PharmacotherapyDB} & 34,412 & 26 & 1,691,829 & \{20.83s, 2.41s\} & 3,307 & 23 & 11,905 & \{2.95s, 0.12s\} & 10.17\\
        \textbf{DrugBank} & 35,103 & 109 & 1,789,976 & \{1011.86s, 11.41s\} & 6,335 & 101 & 184,273 & \{146.98s, 3.56s\} & 13.12\\
        \bottomrule
    \end{tabular}
    \end{adjustbox}
    \label{results-efficiency}
\end{table*}

\subsection{Impact of \sys in supervised settings}
We evaluate the influence of subgraphs derived from retrieved paths on supervised learning for LLMs and GNNs.

\textbf{LLM performance:} We fine-tuned Llama 3.1 8B Instruct using \sys and textual definitions.
As shown in \cref{results-efficiency}, fine-tuned LLMs perform comparably regardless of the training source, indicating their ability to integrate both structured and unstructured knowledge when supervision is provided.
However, we further observe that definitions often outperform \sys, likely because they provide a more direct and semantically rich signal that is easier to learn; in contrast, in the zero-shot setting (\cref{result-zero-shot}), \sys provide more useful relational cues.


\textbf{GNN performance:} We compared GNN models trained on the full knowledge graph (Complete KG) against those trained on the subgraphs constructed from the \sys framework.  \Cref{tab:results-3} shows that with \sys, we achieve comparable accuracy to using the Complete KG despite being approximately 90\% smaller.
This reinforces that smaller, task-specific graphs enhance efficiency without significant performance loss.
For example, on DDInter, EmerGNN achieves nearly identical performance using \sys (F1: 68.85\%) compared to Complete KG (F1: 68.00\%), suggesting that a targeted subgraph retains essential knowledge while significantly improving efficiency.
RGCN benefits from \sys, with F1 increasing from 51.47\% (Complete KG) to 52.12\% (\sys) on DDInter and from 15.49\% to 17.51\% on DrugBank.
This highlights the advantage of a more focused graph structure for such models. This result is consistent in the transductive setting shown in \cref{sec-transd}.

\textbf{Efficiency analysis:} \Cref{results-efficiency} presents a comparison of graph statistics and training times before and after applying \sys for subgraph retrieval, along with the retrieval duration in minutes.
The results clearly show that \sys substantially reduces the number of nodes, relations, and triplets, leading to significantly faster training and lower memory overhead.
For instance, on DrugBank, the number of nodes reduces from 35,103 to 6,335, and triplets from 1,789,976 to 184,273. Correspondingly, EmerGNN’s training time per epoch drops from 1011.86s to 146.98s.. 
A similar trend is observed in DDInter, where subgraph pruning leads to a speedup from 606.24s to 102.21s per epoch.
Note that \sys includes reverse edges during retrieval, effectively doubling the triplet count. Despite this, the total retrieval time remains under 15 minutes for all datasets. 
We see a similar trend in \cref{tab:path_retrieval} for the test data and show the important relations retained by \sys in \cref{app:hetio}.
These results highlight that \sys improves training efficiency without compromising model performance, reinforcing the scalability of our approach. 

\textbf{LLM vs. GNN performance:} Finally, we observe that supervised fine-tuning enables LLMs to outperform GNNs on most datasets, suggesting that supervised LLMs can effectively leverage multiple structured and unstructured modalities and sometimes even surpass GNNs trained solely on relational graphs.

\section{Conclusion}
We present \textit{K-Paths}, a model-agnostic retrieval framework that introduces a diversity-aware adaptation of Yen’s algorithm to extract biologically meaningful, multi-hop reasoning paths from large biomedical KGs.
These paths serve as structured, relational evidence that supports both prediction and interpretability across model types.
Our experiments show that \sys benefits both LLMs and GNNs.
For LLMs, the retrieved paths serve as useful context that enables zero-shot, inductive reasoning about unobserved drug–disease interactions.
For GNNs, \sys identifies compact subgraphs, reducing graph size by up to 90\% while maintaining performance and improving training efficiency.
While our focus is on repurposing and interaction prediction, \sys is potentially generalizable to other biomedical tasks such as protein–protein interaction prediction or treatment recommendation.
We acknowledge limitations: \sys relies on existing interaction types and graph connectivity, which may limit its ability to infer entirely novel interaction types or handle sparsely represented entities.
Nevertheless, by anchoring predictions in relationally diverse paths, \sys mitigates hallucinations and supports biologically grounded inference.
To our knowledge, \sys is one of the first frameworks to enable path-based heuristics for reasoning over unseen drug pairs, a critical need for early-stage drug discovery.
We believe this work lays a foundation for future exploration of integrating LLMs and KGs in biomedical applications, paving the way for more efficient and interpretable solutions in drug discovery and other related fields.

\section*{Acknowledgements}
This material is based in part upon work supported by the National Science Foundation under Grant No. RISE-2425380. Any opinions, findings, and conclusions or recommendations expressed in this material are those of the author(s) and do not necessarily reflect the views of the National Science Foundation. Disclosure: Stephen Bach is an advisor to Snorkel AI, a company that provides software and services for data-centric artificial intelligence.
\bibliographystyle{ACM-Reference-Format}
\bibliography{custom}

\appendix
\newcommand{\cmark}{\textcolor[rgb]{0,0.6,0}{\checkmark}}
\newcommand{\xmark}{\textcolor{red}{$\times$}}

\section{LLM prompting} \label{app:prompting}
We use the following prompt templates across datasets.
As shown below, we provide predefined answer options for the DDInter and PharmacotherapyDB datasets. 
However, we do not include options for the DrugBank dataset for two reasons: 
(1) DrugBank contains 86 possible interaction types, making inference for approximately 30,000 examples computationally expensive.
and (2) preliminary experiments showed that the model performed better without predefined options.
As baselines, we either exclude knowledge graph information entirely or provide textual definitions of the drugs or diseases.
For Tx-Gemma, we followed the original authors' prompt instruction format \cite{wang2025txgemma}. 

\newcommand{\UnifiedPrompt}[5]{
    \begin{tcolorbox}[
        colback=gray!10,
        colframe=black!75!white,
        title=\textbf{Unified prompt template},
        breakable,
        width=\linewidth
    ]
        \textbf{System Prompt:} \\[0.5em]
        You are a pharmacodynamics expert. Answer the question using the given knowledge graph information (if available), essential drug definitions, and your medical expertise. \\
        Base your answer on evidence of known interaction mechanisms, pharmacological effects, or similarities to related compounds, if applicable. Avoid generalizations unless directly supported. \\
        Your answer must be concise and formatted as follows: \#\#Answer:<Dataset Specific>\\
        \textbf{Dataset-Specific Instructions:}
        \begin{itemize}
        \item \textbf{DDInter}: Assess interaction severity \& format as: \textbf{Answer:} \textit{<Major / Moderate / Minor>} 
          
            \item \textbf{PharmacotherapyDB}: Determine therapeutic indication \& format as:  
                  \textbf{Answer:} <\textit{Disease modifying / Palliates / Nonindication}>
            \item \textbf{DrugBank}: Identify mechanism/effect type \& format as:  
                 \textbf{Answer:} <\textit{DrugX mechanism/effect on DrugY}>
        \end{itemize}
        
        \textbf{Question:} \\[0.2em]
        #1 \\
        
        \ifstrempty{#2}{}{
            \textbf{Knowledge Graph Information:} \\[0.2em] <Knowledge graph data>
        }
    \end{tcolorbox}
}

\section*{Prompt template}

\UnifiedPrompt{
    Determine the interaction type or therapeutic indication when (EntityX) and (EntityY) are used together.
}{
}{}{}{}

\section{Qualitative analysis for edge cases}
\label{sec:extra-qualitative}
We inspect two representative queries that highlight the strengths and limitations of \sys.

\textbf{Partial KG coverage.} \\
\textit{Query:} The interaction severity for \textbf{Mipomersen} \& \textbf{Oxymetholone}? (Label: \textsc{Major}) \\
\textit{Paths:} Mipomersen $\rightarrow$ \textsc{major} $\rightarrow$ (Regorafenib, Diclofenac) \\
\textit{Note:} No retrieved evidence for Oxymetholone. \\
\textit{Predictions:} \textbf{Base:} \textsc{Minor}, \textbf{+Def:} \textsc{Major}, \textbf{+\sys:} \textsc{Moderate}

This case shows that while partial KG coverage offers helpful cues (e.g., Mipomersen’s links to severe interactions), models may underutilize these cues without additional context.

\vspace{0.5em}
\textbf{No KG coverage.} \\
\textit{Query:} An example where no paths exist for either entity. \\
In such cases, \sys defaults to the Base LLM’s parametric knowledge. We observe mixed results: Base sometimes predicts correctly, and definitions offer occasional gains. This exposes a natural ceiling for any retrieval-based approach when the external source is unavailable.


\section{Textual definitions with \sys} 
We explored using definitions with \sys in zero-shot settings. It leads to longer input contexts and does not consistently improve performance. For instance, on PharmacotherapyDB with Llama-3.1-70B, the F1 score slightly decreases from 67.46 (\sys) to 66.00 (\sys + Descriptions); on DDInter, performance remains unchanged (46.19 to 46.18). The added context also increases inference time and cost, making fine-tuning expensive. Based on these findings, we focus on path-only retrieval for scalable and cost-effective use.

\section{QLORA fine-tuning} \label{app:qlora}
For the supervised LLM experiments, we fine-tuned Llama 3.1 8B Instruct using QLoRA. We conducted training experiments under two distinct scenarios across our datasets.
In the first scenario, we trained the model using the retrieved paths for each training query.  
In the second scenario, we trained the model using text definitions of the drugs or diseases. 
Definitions were capped at 200 tokens, reflecting the dataset's average definition length.
In both scenarios, we trained for 10 epochs using the default settings of the QLoRA repository, with the following modifications: 
A learning rate of 1e-3 and the maximum input length to the average token length of the input across the respective dataset. 
The training was conducted on 8 A100-80G GPUs and typically completed within 24 hours, depending on the dataset.
During inference, we first retrieved reasoning paths using \sys.
These retrieved paths were then appended to the original query and fed into the fine-tuned LLM to generate the final answers.

\section{GNN baselines}
\subsection{Relational graph convolutional network}
We implement the Relational Graph Convolutional Network (RGCN) \cite{Schlichtkrull2017ModelingRD}, which operates on the augmented graph with multiple relation types and employs message passing to propagate structured information across nodes. The model is implemented using PyTorch and PyTorch Geometric.

\textbf{Node feature initialization:} Drug and disease nodes are initialized using RoBERTa \cite{liu2019robertarobustlyoptimizedbert} embeddings extracted from PubMed-scraped descriptions.
Other entity nodes (genes, anatomy, etc.) are initialized randomly, allowing the model to learn meaningful representations during training.


\textbf{Training setup:} We follow the inductive setting for dataset splitting, as described in \cref{sec:datasets}. 
The training follows a link prediction framework where training nodes are sampled with all their relations observed, while test nodes are introduced to evaluate generalization. We consider two training settings:
(1) Training on the entire augmented KG (\textit{Complete KG}) and testing on test nodes along with their retrieved test KG, and (2) Training on the diverse retrieved train paths and testing on test nodes along with their retrieved test paths.

\textbf{Model architecture \& training details:} We train a three-layer RGCN using the Adam optimizer with a learning rate of 1e-3 and a scheduler based on validation loss. Cross-entropy loss is used to predict drug–drug or drug–disease interactions. Training runs for up to 1,000 epochs with early stopping. To ensure class balance, stratified sampling selects up to 1,000 samples per epoch and 10 per class. Each GCN layer is followed by ReLU, projecting node embeddings into a 128-dimensional space with batch normalization, ReLU, and dropout (rate 0.5). An edge classifier predicts interaction types using the final embeddings of entity pairs. To improve efficiency, we use basis decomposition with two bases.

\subsection{EmerGNN}

To compare against RGCN, we evaluate EmerGNN, a graph neural network designed for emerging drug-drug interaction prediction \cite{zhang2023emergingdruginteractionprediction}. We use the official implementation and apply it to our datasets without modifying the model architecture or training pipeline. Unlike RGCN, which relies on RoBERTa embeddings for node initialization, EmerGNN incorporates molecular features, leveraging structural and chemical properties to enhance node representation.
We compare the performance of both models in terms of interaction prediction accuracy, assessing the impact of different node initialization strategies and augmented KG utilization.

\section{Additional experimental results}
\label{sec:additional_results}

This section presents additional experimental results, including retrieved paths from the augmented KG, dataset statistics, path retrieval efficiency, and model performance comparisons.

\subsection{Dataset overview}
\Cref{tab:datasets} summarizes the datasets used in our experiments, categorized by prediction task and the connectivity between interaction query nodes (entities) in the augmented KG.

\begin{itemize}
    \item DrugBank involves inductive and transductive tasks, predicting drug-drug interactions among 86 labels. The transductive setting has more drug pairs connected in the augmented KG (38,411) than the inductive setting (27,983).
    \item DDInter predicts drug-drug interaction severity levels (Major, Moderate, or Minor). It contains 13,841 connecting drug pairs, and 5,494 interaction queries contain information about a single entity.
    \item PharmacotherapyDB focuses on whether a drug is disease-modifying, palliates, or has no indication of a disease.
\end{itemize}

\begin{table}[t]
    \caption{\small Summary of datasets and tasks.}
    \centering
    \footnotesize 
    \setlength{\tabcolsep}{2pt} 
    \renewcommand{\arraystretch}{1.1} 
    \begin{tabular}{@{}lp{1.8cm}rrr@{}}
        \toprule
        \textbf{Dataset} & \textbf{Task} & \textbf{Two Nodes} & \textbf{Single Nodes} & \textbf{No Node} \\ 
        \midrule
        DrugBank (Ind.) & Open-ended & 27,983 & 3,987 & 14 \\ 
        DrugBank (Trans.) & Open-ended & 38,411 & 8 & 0 \\ 
        DDInter & Categorical & 13,841 & 5,494 & 104 \\ 
        PharmacotherapyDB & Categorical & 252 & 0 & 0 \\ 
        \bottomrule
    \end{tabular}
    \label{tab:datasets}
\end{table}

\begin{table}[t]
    \caption{\small Performance of models on the Transductive DrugBank setting. Bold indicates the best performance, and underlined denotes the second-best. LLMs leverage both knowledge types effectively with supervision, and \sys enhance GNN efficiency
    without significant performance loss.}
    \centering
    \footnotesize
    \renewcommand{\arraystretch}{0.9} 
    \setlength{\tabcolsep}{4pt} 
    \begin{tabular}{llccc}
        \toprule
        \textbf{Model} & \textbf{Setting} & \textbf{Accuracy} & \textbf{F1} & \textbf{Kappa} \\
        \midrule
        \multicolumn{5}{l}{\textbf{Graph-Based Models}} \\
        EmerGNN & Complete KG & \textbf{97.40} & \underline{94.00} & \underline{96.60} \\
        EmerGNN & \sys & \underline{97.01} & \textbf{94.25} & \textbf{97.01} \\
        RGCN & Complete KG & 90.01 & 87.62 & 89.85 \\
        RGCN & \sys & 90.86 & 88.43 & 90.11 \\
        SumGNN & Reported & 86.85 & 92.66 & 92.66 \\
        KnowDDI & Reported & 91.53 & 93.17 & 91.89 \\
        Decagon & Reported & 87.20 & 57.40 & 86.10 \\
        \midrule
        \multicolumn{5}{l}{\textbf{LLM-Based Models}} \\
        QLoRA-Llama & Definitions & 93.45 & 91.71 & 92.28 \\
        QLoRA-Llama & \sys & 93.58 & 88.98 & 92.41 \\
        \bottomrule
    \end{tabular}
    \label{tab:results-transductive}
\end{table}

\begin{table}[h]
    \caption{\small Statistics: Comparison of the Augmented KG with extracted subgraph at test time. Retrieval time reflects the cost of extracting \sys for all test queries.}
    \centering
    \begin{adjustbox}{max width=0.5\textwidth}
    \begin{tabular}{lccccccc}
        \toprule
        \multirow{2}{*}{Dataset} & \multicolumn{3}{c}{Before Retrieval (Augmented KG)} & \multicolumn{4}{c}{After Retrieval (\sys)} \\
        \cmidrule(lr){2-4} \cmidrule(lr){5-8}
         & \#Nodes & \#Relations & \#Triplets & \#Nodes & \#Relations & \#Triplets & Time (min)\\
        \midrule
        DrugBank & 35,146 & 102 & 1,722,677 & 6,378 & 94 & 175,698 & 13.00 \\
        DDInter  & 35,169 & 26  & 1,710,079 & 5,647 & 18 & 102,854  & 18.01 \\
        PharmDB  & 33,952 & 26  & 1,690,945 & 2,847 & 23 & 13,942   & 10.15 \\
        \bottomrule
    \end{tabular}
    \end{adjustbox}
    \label{tab:path_retrieval}
\end{table}

\begin{table}[t]
    \caption{\small Hetionet relations retained by \sys. Relation names use: A = Anatomy, D = Disease, G = Gene, C = Compound, SE = Side Effect, PC = Pharmacologic Class, BP = Biological Process, CC = Cellular Component, MF = Molecular Function, PW = Pathway. \cmark = present, \xmark = absent.
    }
    \centering
     \footnotesize
    \renewcommand{\arraystretch}{1.0}
    \setlength{\tabcolsep}{2pt}
    \begin{tabular}{@{}p{0.3cm}p{2.5cm}>{\centering\arraybackslash}p{1.5cm}>{\centering\arraybackslash}p{1.5cm}>{\centering\arraybackslash}p{1.5cm}@{}}
        \toprule
        \textbf{ID} & \textbf{Relation} & \textbf{Drugbank} & \textbf{DDinter} & \textbf{PharmDB} \\
        \midrule
        0  & A--downregulates--G & \xmark & \xmark & \cmark \\
        1  & A--expresses--G & \xmark & \xmark & \cmark \\
        2  & A--upregulates--G & \xmark & \xmark & \cmark \\
        3  & C--binds--G & \cmark & \cmark & \cmark \\
        4  & C--causes--SE & \cmark & \cmark & \cmark \\
        5  & C--downregulates--G & \cmark & \cmark & \cmark \\
        6  & C--palliates--D & \cmark & \cmark & \cmark \\
        7  & C--resembles--C & \cmark & \cmark & \cmark \\
        8  & C--treats--D  & \cmark & \cmark & \cmark \\
        9  & C--upregulates--G  & \cmark & \cmark & \cmark \\
        10 & D--associates--G & \cmark & \cmark & \cmark \\
        11 & D--downregulates--G & \cmark & \cmark & \cmark \\
        12 & D--localizes--A & \xmark & \xmark & \cmark \\
        13 & D--presents--SE & \xmark & \xmark & \cmark \\
        14 & D--resembles--D & \cmark & \cmark & \cmark \\
        15 & D--upregulates--G & \cmark & \cmark & \cmark \\
        16 & G--covaries--G & \cmark & \cmark & \cmark \\
        17 & G--interacts--G & \cmark & \cmark & \cmark \\
        18 & G--participates--BP & \xmark & \xmark & \xmark \\
        19 & G--participates--CC & \xmark & \xmark & \xmark \\
        20 & G--participates--MF & \xmark & \xmark & \xmark \\
        21 & G--participates--PW & \xmark & \xmark & \xmark \\
        22 & G--regulates--G & \cmark & \cmark & \cmark \\
        23 & PC--includes--C  & \cmark & \cmark & \cmark \\
        \bottomrule
    \end{tabular}
    \label{tab:relations_pruning}
\end{table}

\subsection{Transductive results}\label{sec-transd}
\Cref{tab:results-transductive} compares the performance of GNNs and LLM-based models on the Drugbank transductive dataset. Among GNNs, EmerGNN performs best, achieving 97.40\% accuracy with the complete KG, while using \sys slightly lowers accuracy (97.01\%) but improves the F1-score. 
RGCN performs worse than EmerGNN but benefits from our KG, increasing accuracy from 90.01\% to 90.86\%. 
For LLM-based models, QLoRA-Llama achieves 93.58\% accuracy when using \sys, while using text-based descriptions instead of structured knowledge results in similar accuracy (93.45\%) but a slightly higher F1-score. 
Overall, EmerGNN performs best and \sys improves efficiency without significant performance loss.

\subsection{Augmented KG vs. Test-Time subgraph}

\Cref{tab:path_retrieval} compares the augmented KG's overall structure to the subgraph extracted at test time. 
We report the number of nodes, relations, and triplets before and after query-specific retrieval. 
Filtering for relevant subgraphs significantly reduces graph size: 
DrugBank shrinks from 1.7M to 175K triplets, DDInter from 1.71M to 102K, and PharmacotherapyDB from 1.69M to 13K. 
This demonstrates the efficiency of query-specific retrieval in extracting only the most relevant paths for inference.

\subsection{Retained Hetionet relations}\label{app:hetio}

\Cref{tab:relations_pruning} presents the Hetionet relations retained in the training and test subgraphs after \sys was applied to each dataset. These relations encapsulate biological interactions, such as gene participation in molecular functions or compounds treating diseases.

Since Hetionet serves as a structured biomedical KG, these relations may not be inherent to the datasets (DrugBank, DDInter, PharmacotherapyDB) themselves 
but instead provide additional contextual knowledge that enhances reasoning within the models. The path retrieval process selectively retains the most informative relations while discarding those less relevant to each dataset.

For example, in the DDInter dataset: 
\begin{itemize} 
    \item Hetionet originally contained 23 distinct relation types, covering diverse biological interactions. 
    \item After dataset-specific path extraction, only 15 relation types were retained by \sys and used for training, ensuring that only the most relevant interactions contributed to the model. 
\end{itemize}

Interestingly, while PharmacotherapyDB is the smallest dataset, it retains more relations than DrugBank and DDInter. This is because PharmacotherapyDB encompasses both drug and disease entities, covering a broader set of biomedical interactions that span multiple entities.

\end{document}